\documentclass[letterpaper, 10 pt, conference]{ieeeconf}  

\IEEEoverridecommandlockouts                              

\usepackage{graphicx} 

\usepackage{booktabs}
\usepackage{multirow}
\usepackage{subcaption}
\usepackage{hyperref}
\usepackage{url}
\usepackage{xcolor}

\usepackage{placeins}

\title{\LARGE \bf
Improving 6D Object Pose Estimation of metallic Household and Industry Objects
}

\usepackage[style=ieee,citestyle=numeric-comp]{biblatex}
\author{Thomas Pöllabauer$^{1}$, Michael Gasser$^{1,2}$, Tristan Wirth$^{1}$, Sarah Berkei$^{3}$, Volker Knauthe$^{1}$, Arjan Kuijper$^{1,2}$
\thanks{$^{1}$Technical University Darmstadt, Germany
        }%
\thanks{$^{2}$Fraunhofer Institute for Computer Graphics Research IGD, Germany}%
\thanks{$^{3}$Threedy GmbH, Germany}%
}

\begin{document}

\maketitle
\thispagestyle{empty}
\pagestyle{empty}

\begin{abstract}

6D object pose estimation suffers from reduced accuracy when applied to metallic objects. We set out to improve the state-of-the-art by addressing challenges such as reflections and specular highlights in industrial applications. Our novel BOP-compatible dataset \cite{32,hodan2024bop}, featuring a diverse set of metallic objects (cans, household, and industrial items) under various lighting and background conditions, provides additional geometric and visual cues. We demonstrate that these cues can be effectively leveraged to enhance overall performance. To illustrate the usefulness of the additional features, we improve upon the GDRNPP \cite{liuShanicelGdrnpp_bop20222024} algorithm by introducing an additional keypoint prediction and material estimator head in order to improve spatial scene understanding. Evaluations on the new dataset show improved accuracy for metallic objects, supporting the hypothesis that additional geometric and visual cues can improve learning.

\end{abstract}
  
\section{INTRODUCTION}
Estimating the six degrees of freedom (6D) pose of an object in RGB or RGB-D images is a fundamental problem in computer vision with broad applications, including robotics, autonomous driving, and augmented reality. Robots require precise pose estimation to manipulate objects, while self-driving cars depend on it for obstacle detection and navigation. Similarly, augmented reality systems rely on accurate pose estimation to seamlessly integrate virtual objects into real-world environments.

Despite significant advancements in deep learning-based 6D pose estimation, challenges persist; particularly when dealing with metallic objects. The reflective and specular nature of metallic surfaces leads to complex issues such as lighting-induced distortions, environmental reflections, and occlusions, resulting in reduced estimation accuracy. Recent benchmarks, such as the BOP challenge, have highlighted the reduced performance of existing models on metallic objects compared to non-metallic surfaces, underscoring the need for specialized solutions.

This work extends the GDR-Net/GDRNPP \cite{wang2021gdr,liuShanicelGdrnpp_bop20222024} architecture to improve 6D pose estimation for metallic objects. Our key contributions include: (1) a novel \textit{keypoint} generation and heatmap learning strategy to enhance geometric understanding, (2) a material property learning module to mitigate the impact of reflections and specular highlights, and (3) the development of a custom physically-based rendering (PBR) dataset tailored for evaluating estimation quality on metallic objects. These enhancements enable more robust and accurate pose estimation in real-world industrial and robotic applications.

\section{RELATED WORK}
\subsection{6D Pose Estimation}
There are multiple approaches that address the estimation of a pose of an object relative to the camera within six degrees of freedom, i.e., rotation and translation, based on RGB or RGBD images.
He et al. \cite{6} distinguish between non-learning-based methods such as Linemod \cite{hinterstoissermultimodal}, that utilize traditional computational techniques to estimate an object's pose, and learning-based methods, that employ data-driven learning.
Learning-based strategies have shown a high level of accuracy in recent years.
Liu et al. \cite{23} further subdivide learning-based approaches into \textit{category-level} strategies, that generalization to objects within a category, \textit{instance-level} methods, which estimate the 6D pose of certain known objects, and approaches, that perform 6D pose estimation on \textit{unseen objects}.
We propose a novel approach to \textit{instance-level} 6D pose estimation of known metallic household and industry objects.

\subsection{GDRNPP}
This work builds on GDRNPP \cite{liuShanicelGdrnpp_bop20222024}, which constitutes a high-performing 6D object pose estimation algorithm that builds on GDR-Net \cite{wang2021gdr} exchanging its backbone with ConvNeXt. We choose this method for it's simple and easy-to-extend design, as demonstrated in previous modifications \cite{pollabauer2024extending,poellabaer2024epro}, and high performance in the BOP challenge \cite{bop_leaderboard}.
GDRNPP takes an RGB image as input and detects relevant image regions, that are likely to contain the wanted object in an initial step.
A convolutional neural network (CNN) consisting of a backbone architecture and a multi-headed decoder (GeoHead) extracts relevant feature maps from these relevant image regions.
Subsequently, a Patch-PnP module directly regresses the rotation and translation from these feature maps.
As a final step, there is an optional depth refinement step.

In this work, we focus on extending the output of the GeoHead part to predict additional geometric features and material characteristics.

\subsection{Datasets and Benchmarks}
6D pose estimation is required across multiple domains with diverse environmental conditions from household and outdoor settings to industrial environments.
The BOP challenge provides a framework to benchmark 6D pose estimation strategies in a comparable manner, encompassing diverse datasets, including textured \cite{37,41,42} and textureless \cite{9,40} objects, diverse lighting conditions \cite{32} as well as specialized datasets for industrial applications \cite{9,38}.
However there is no dataset in the BOP challenge, that accurately depicts the adverse nature of metallic surfaces, such as reflections and specular highlights, that appear in diverse settings. The most similar dataset ITODD \cite{8} shows industrial objects, but only in a constrained setting.

\subsection{Bottleneck Attention}
Traditional CNN architectures often have problems capturing long-range dependencies in image space.
Vision Transformers \cite{vit} utilize the attention mechanism established in natural language processing to calculate spatial dependencies in image space.
This however, does not capture the relevance of different channels, which can be numerous in CNNs.
Park et al. \cite{35} adress this issue by proposing Bottleneck Attention Modules (BAM).
BAMs combine seperate spatial and channel-wise attention maps to enhance a model's ability to focus on the relevant information in multi-channel feature maps.


\begin{figure}
    \centering
    \begin{subfigure}{\linewidth}
        \centering
        \includegraphics[width=1\linewidth]{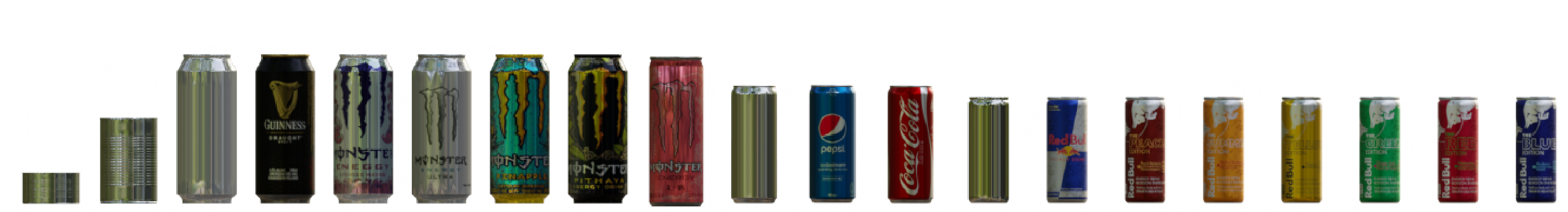}
        \caption{Cans}
        \label{fig:objects:cans}
    \end{subfigure}
    
    \vspace{0.0001cm} 

    \begin{subfigure}{\linewidth}
        \centering
        \includegraphics[width=1\linewidth]{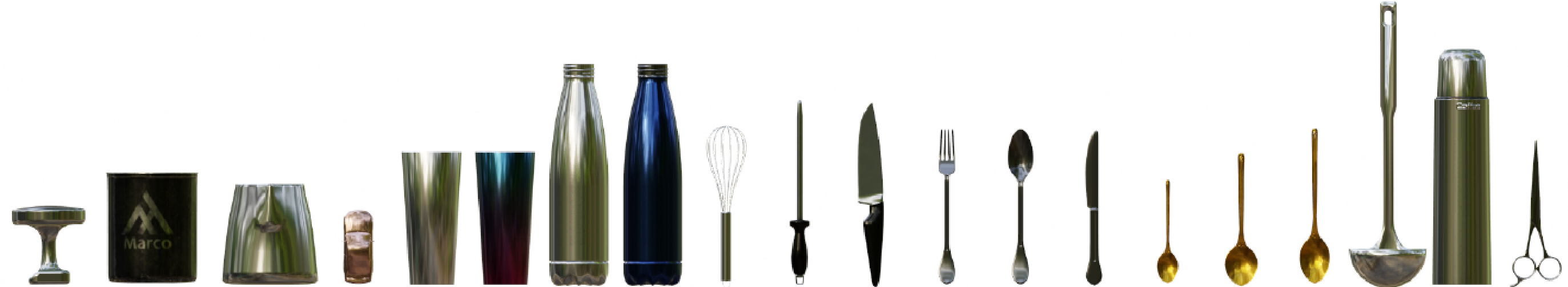}
        \caption{Household}
        \label{fig:objects:household}
    \end{subfigure}

    \vspace{0.1cm} 

    \begin{subfigure}{\linewidth}
        \centering
        \includegraphics[width=1\linewidth]{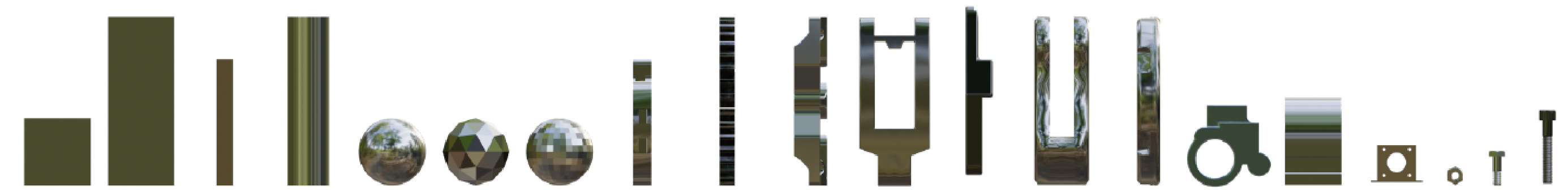}
        \caption{Industry}
        \label{fig:objects:industry}
    \end{subfigure}
    
    \caption{Metallic objects from households and industry that are used in our novel 6D pose estimation dataset.}
    \label{fig:objects}
\end{figure}


\section{METHODOLOGY}

\begin{figure}[t]
    \centering
    \begin{subfigure}{0.33\linewidth}
        \centering
        \includegraphics[width=\linewidth]{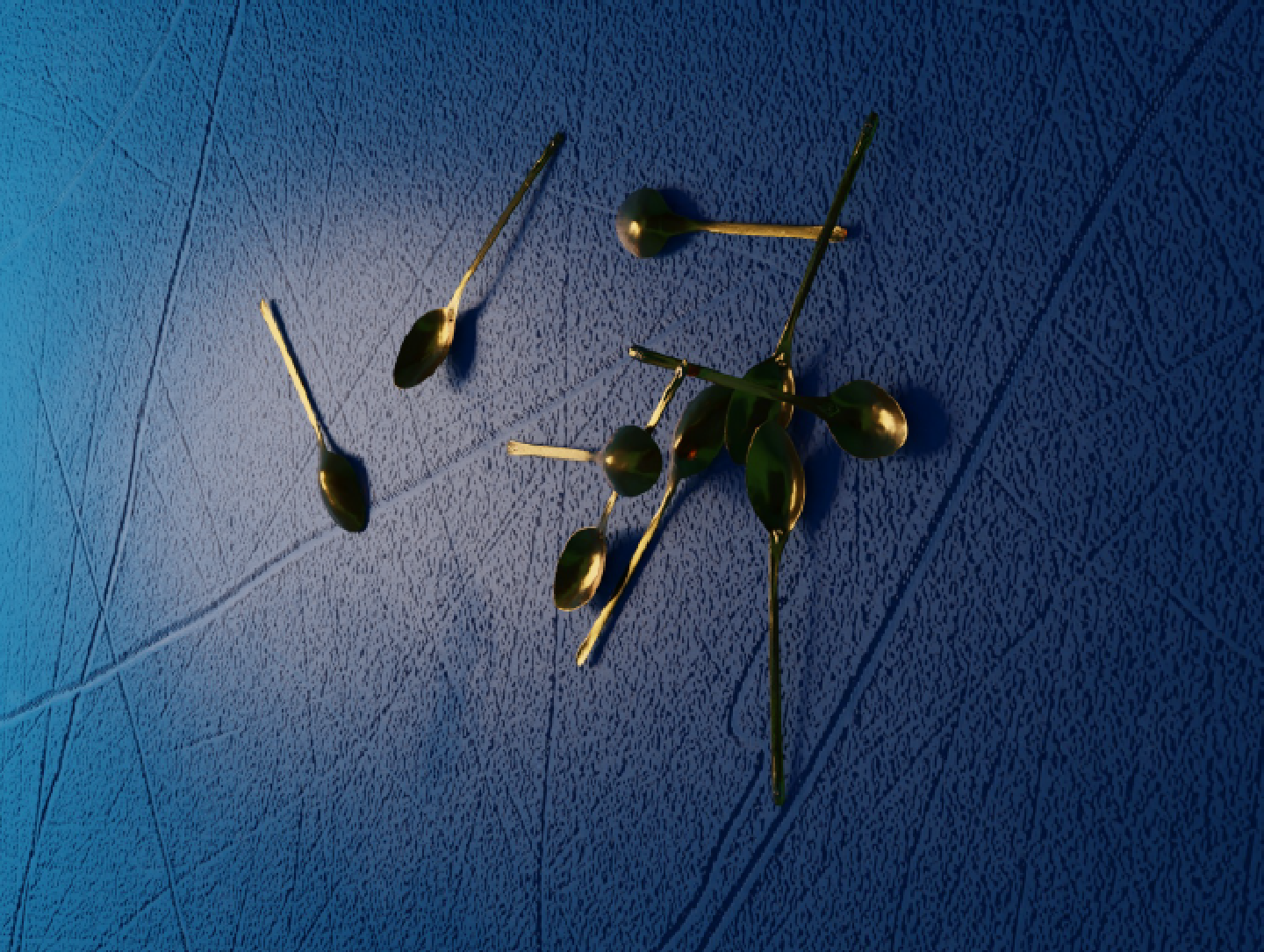}
        \caption{Ambient + Point}
    \end{subfigure}
    \begin{subfigure}{0.33\linewidth}
        \centering
        \includegraphics[width=\linewidth]{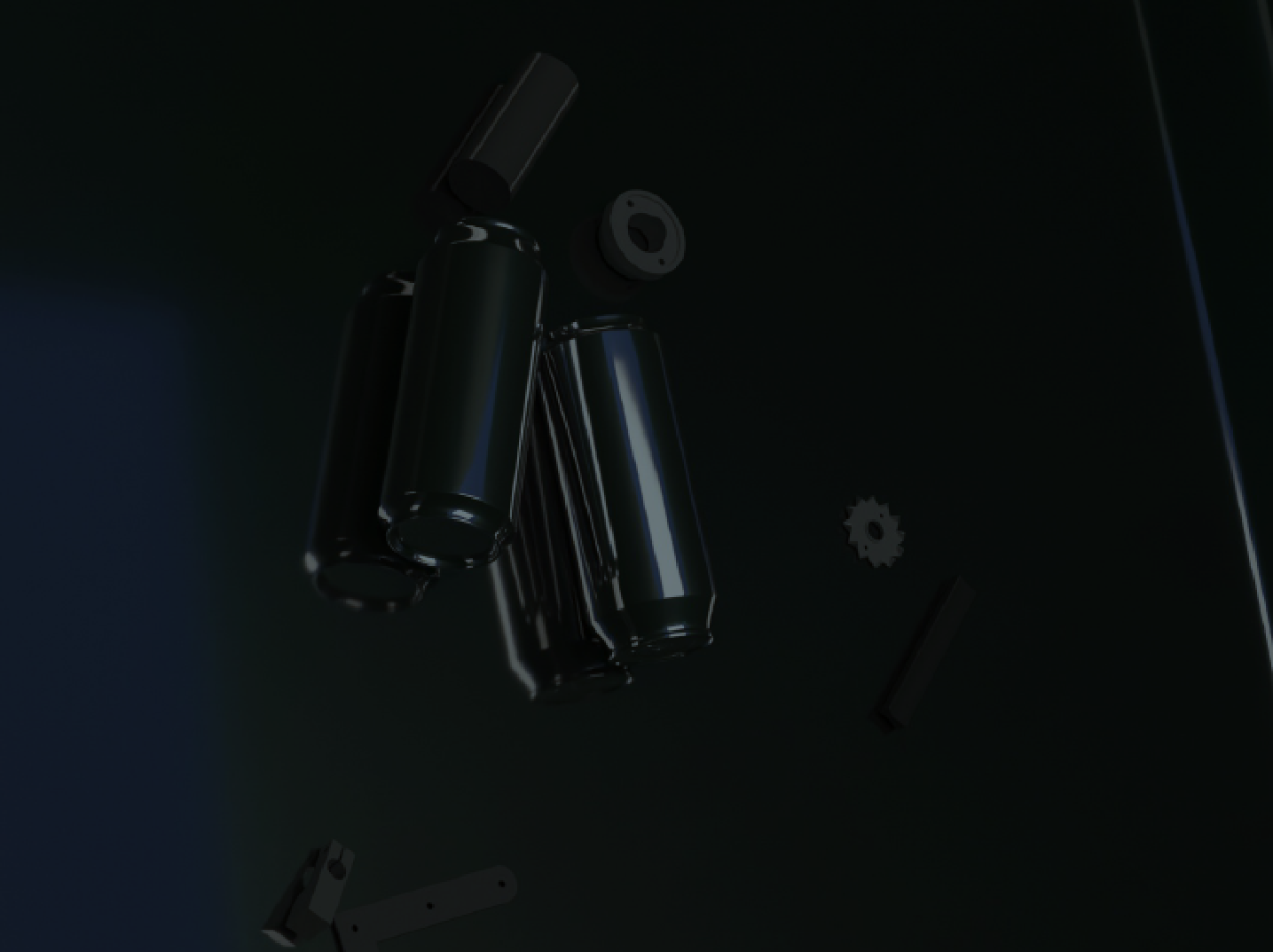}
        \caption{Only Point}
    \end{subfigure}
    
    \vspace{0.1cm} 
    
    \begin{subfigure}{0.32\linewidth}
        \centering
        \includegraphics[width=\linewidth]{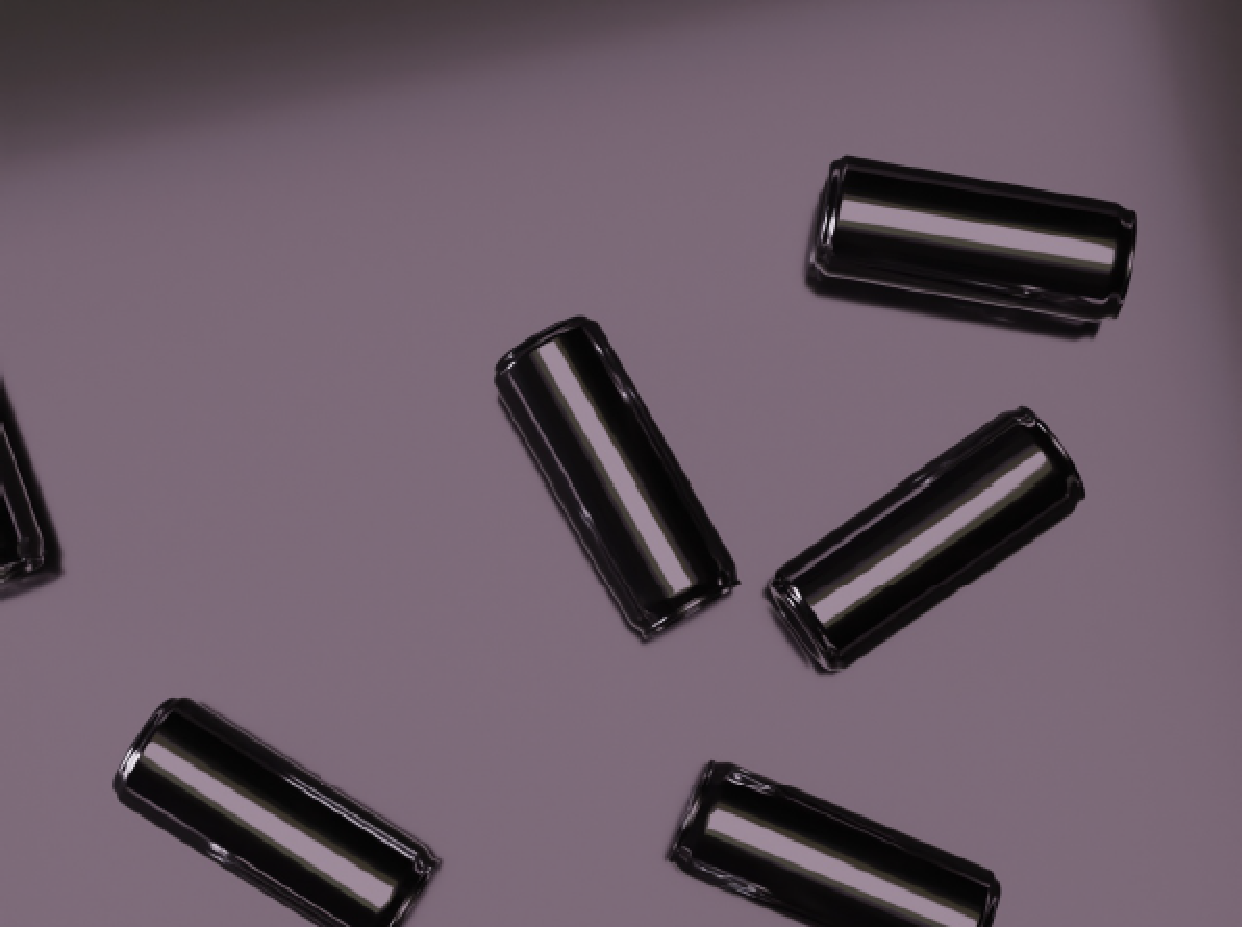}
        \caption{Only Ambient}
    \end{subfigure}
    \begin{subfigure}{0.32\linewidth}
        \centering
        \includegraphics[width=\linewidth]{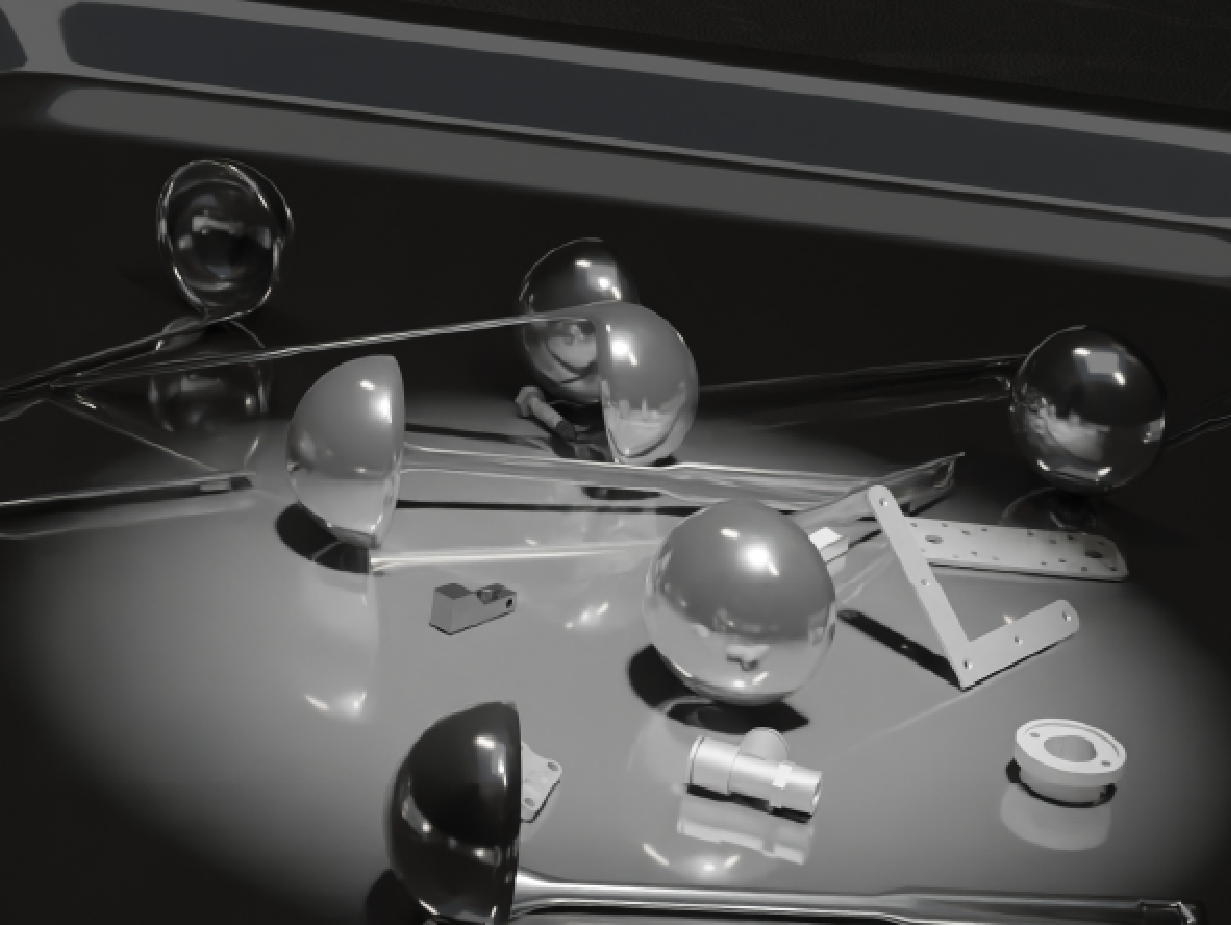}
        \caption{Ambient + Spot}
    \end{subfigure}
    \begin{subfigure}{0.32\linewidth}
        \centering
        \includegraphics[width=\linewidth]{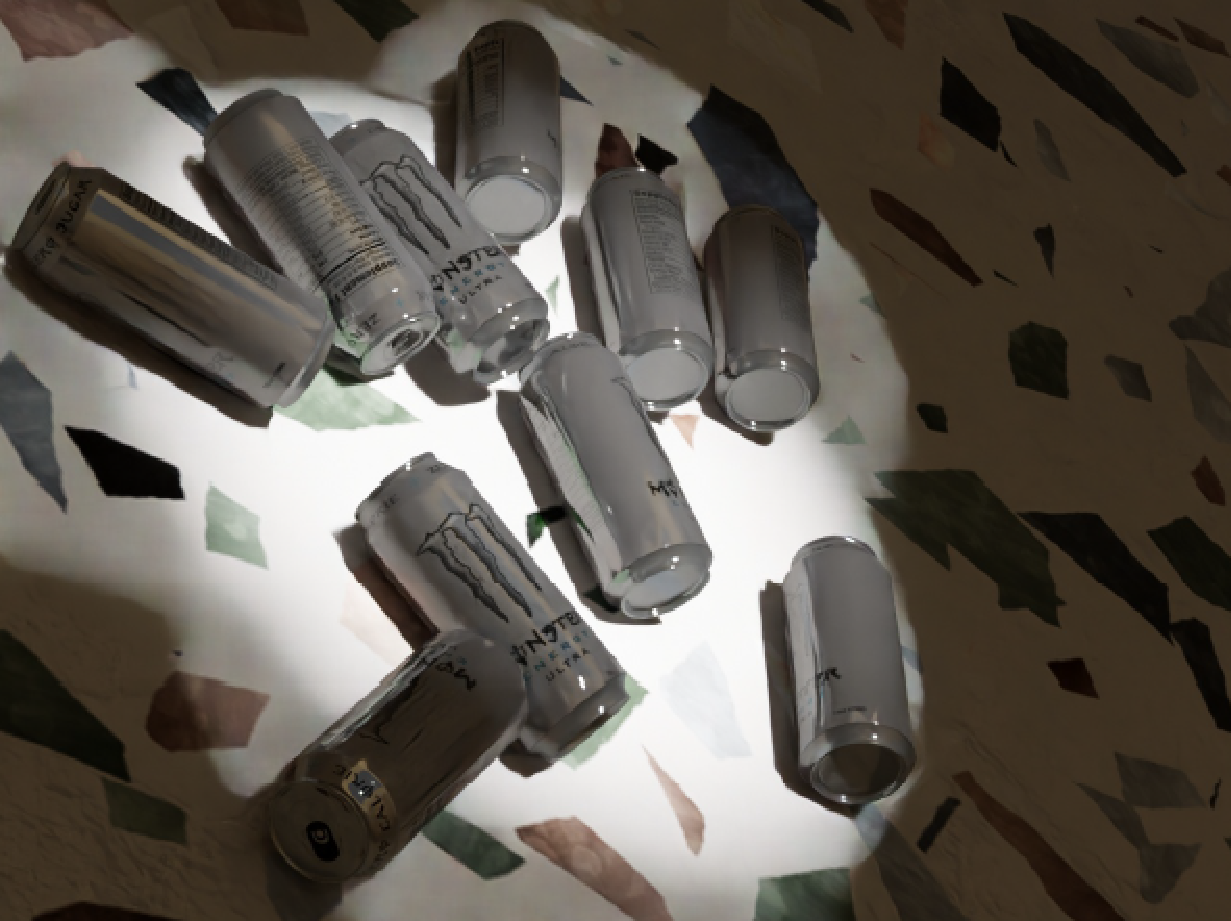}
        \caption{Multiple Spots}
    \end{subfigure}
    
    \caption{Example Scenes from our dataset representing the five lighting scenarios utilizing built-in blender light sources: (a) Ambient light with one point light source, (b) one point light source, (c) only ambient illumination, (d) ambient illumination with a spot light source, and (e) multiple spot light sources.}
    \label{fig:light}
\end{figure}

\begin{figure}[t]
    \centering
    \begin{subfigure}{0.32\linewidth}
        \centering
        \includegraphics[width=\linewidth]{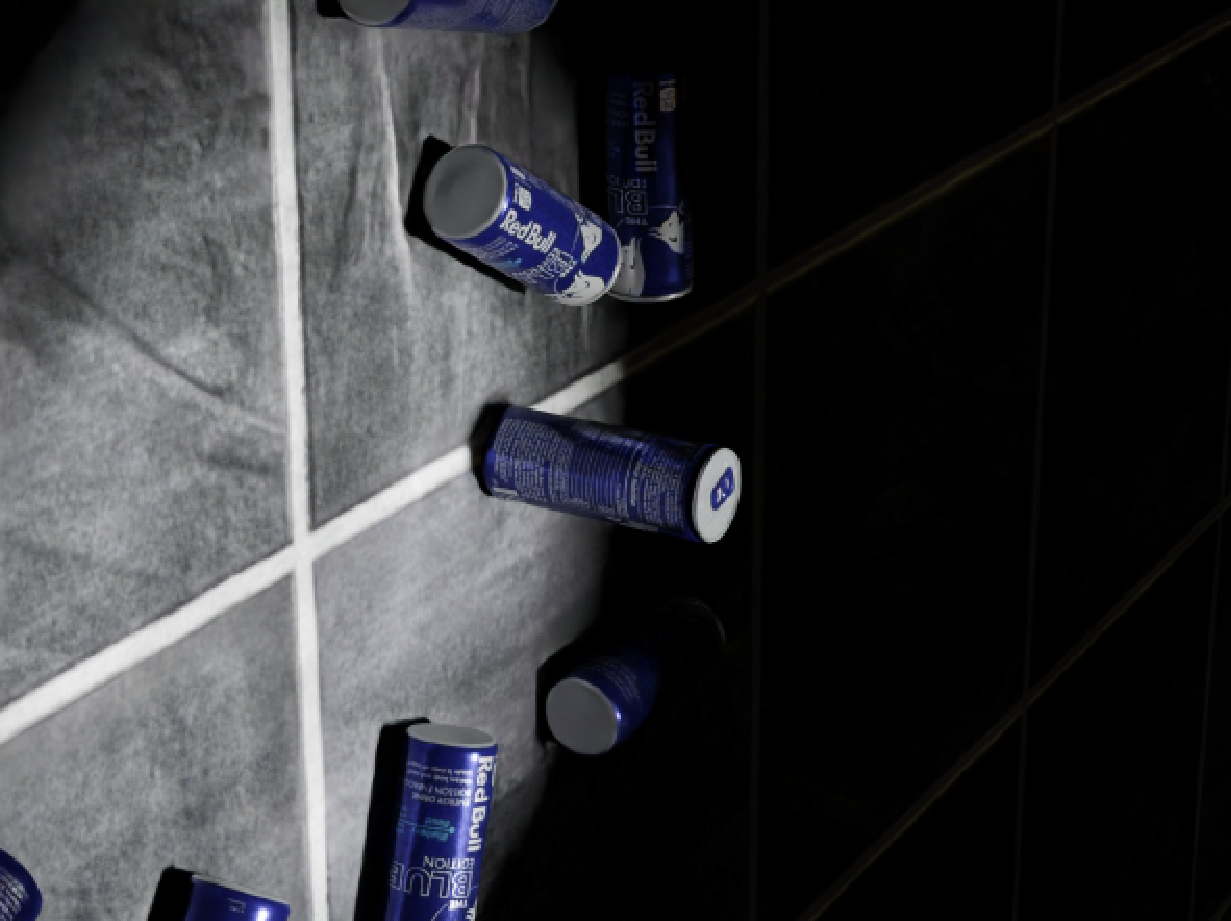}
        \caption{Sample Scene}
    \end{subfigure}
    \hfill
    \begin{subfigure}{0.32\linewidth}
        \centering
        \includegraphics[width=\linewidth]{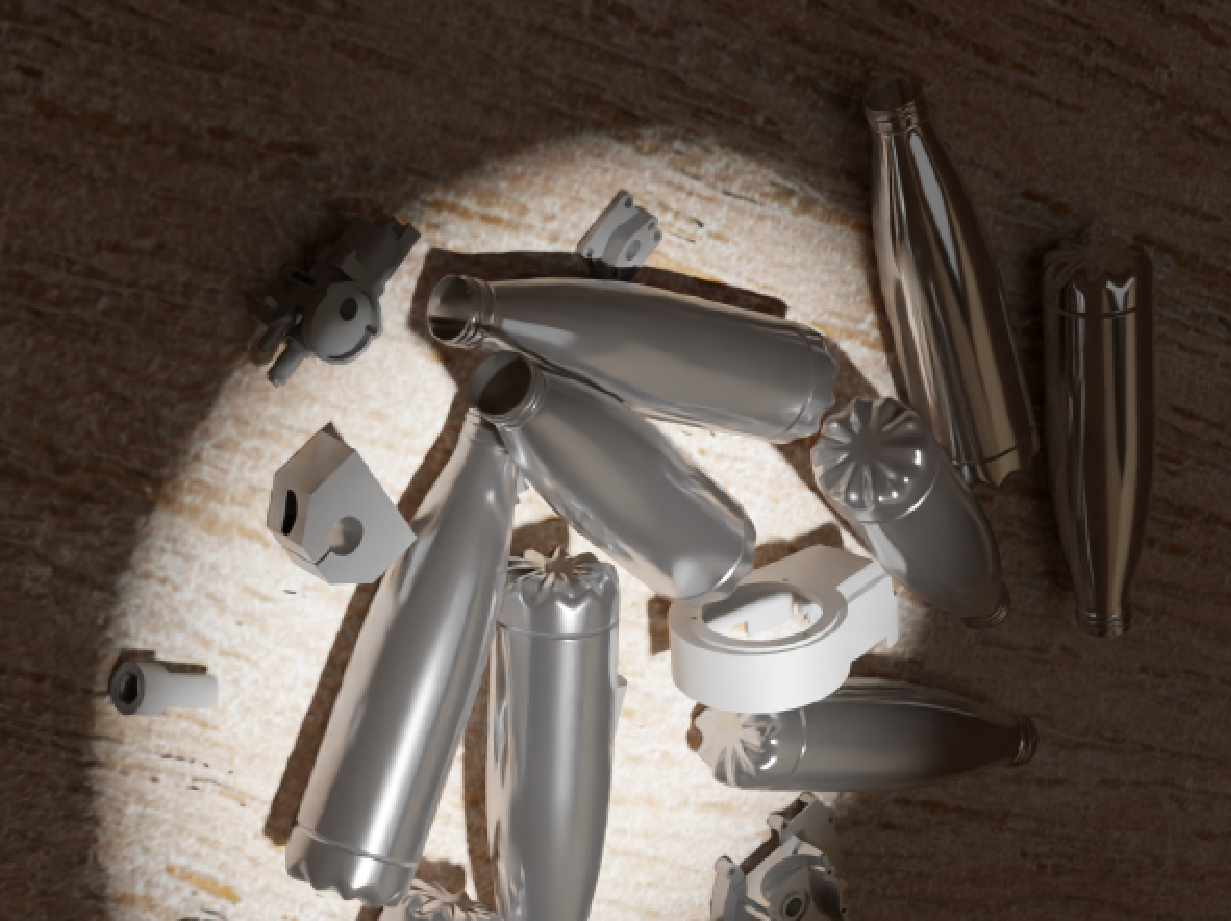}
        \caption{Occluding objects}
    \end{subfigure}
    \hfill
    \begin{subfigure}{0.32\linewidth}
        \centering
        \includegraphics[width=\linewidth]{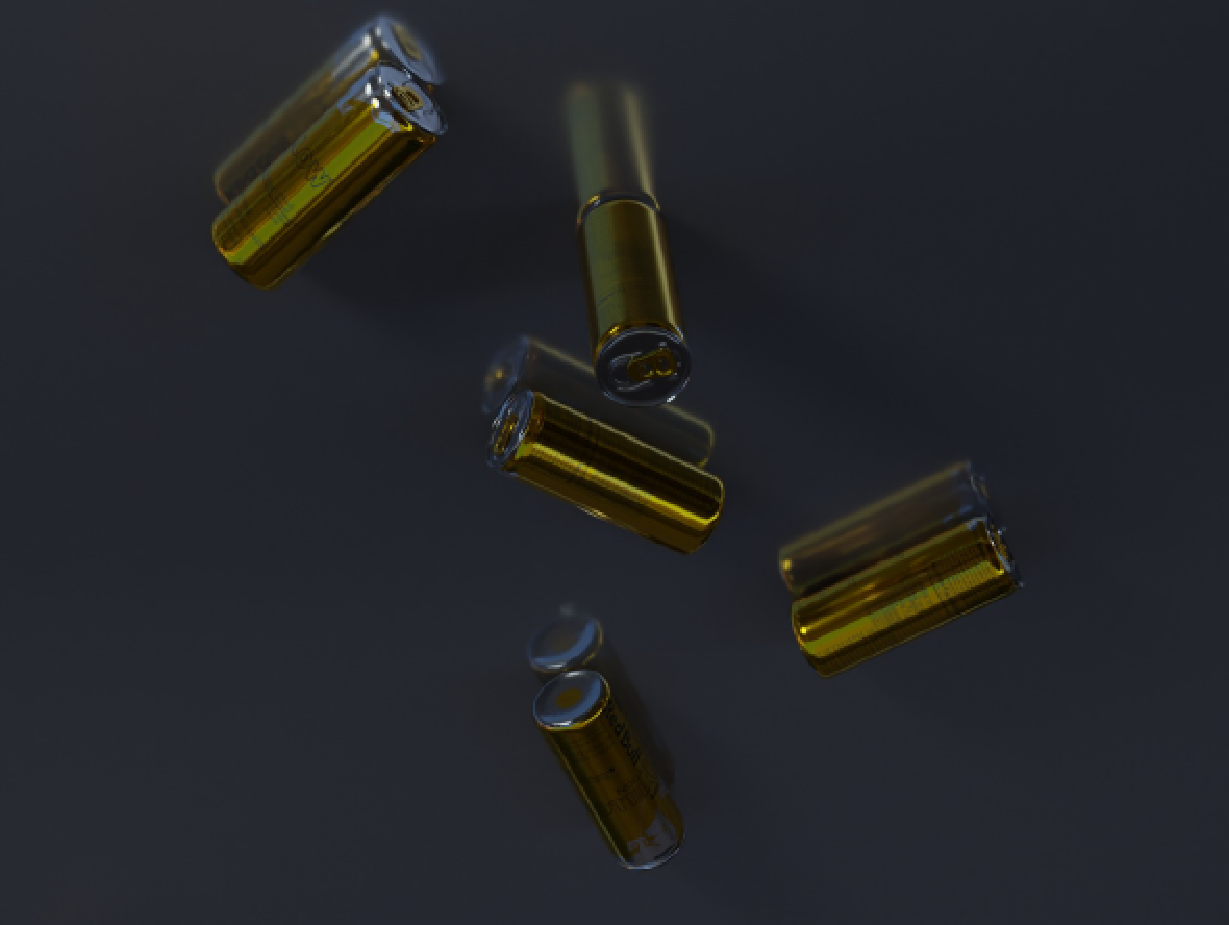}
        \caption{Reflective surface}
    \end{subfigure}
    
    \caption{Images from our dataset (a) without additional modalities, (b) with additional occluding objects from the ITODD dataset, and (c) with reflective surfaces.}
    \label{fig:scenes2}
\end{figure}

\begin{figure*}[t]
    \centering
    \includegraphics[width=0.9\linewidth]{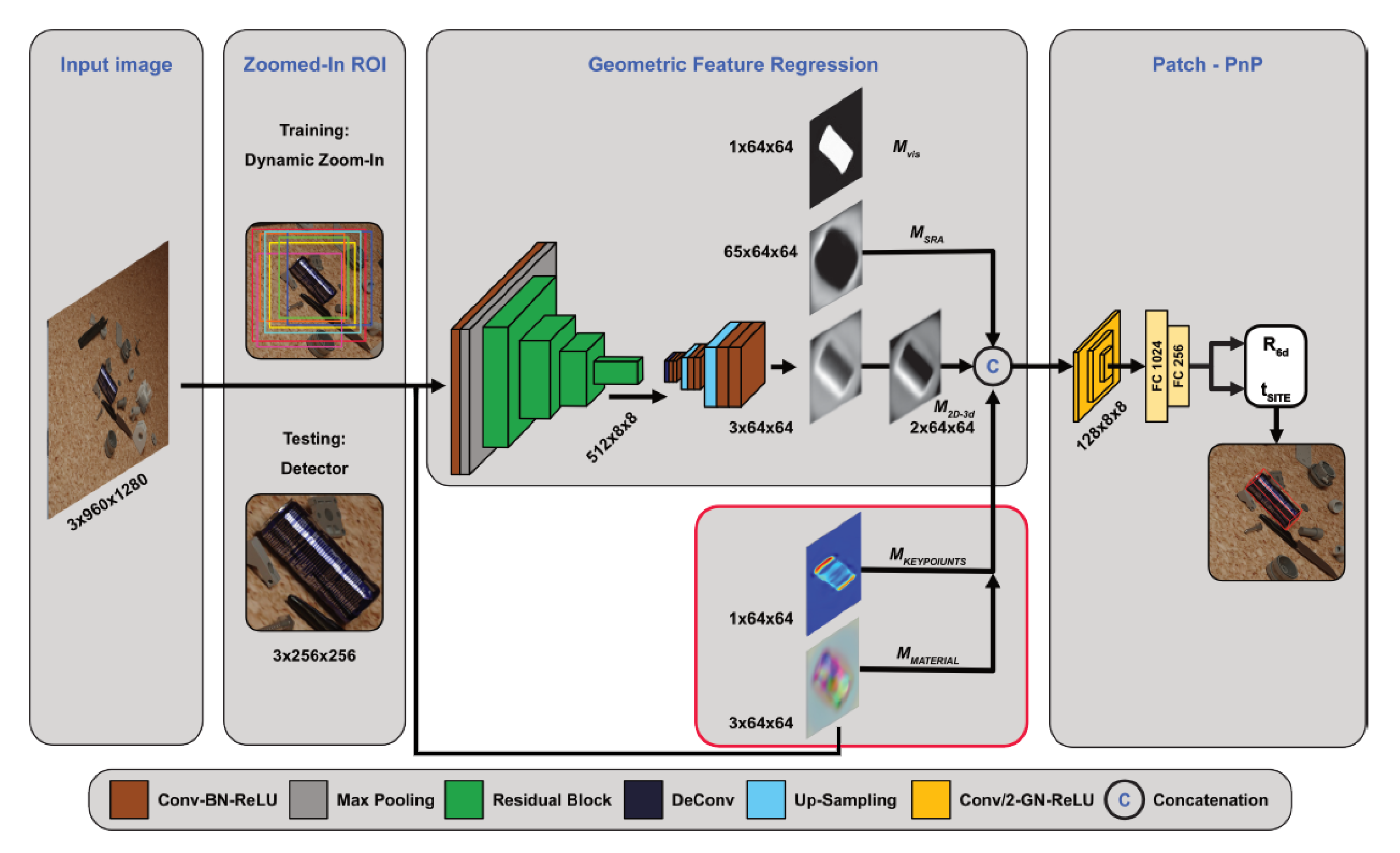}
    \caption{Extension of GDRNPP to predict additional geometric \textit{keypoints} and object material information. Two additional heads in the decoder predict either a heatmap, encoding the projected \textit{keypoint} location or a pixel-wise difference image, representing the necessary change to the input crop, to simulate a non-metallic surface.}
    \label{fig:method}
\end{figure*}

\subsection{Metallic Benchmark Dataset}
We propose a novel BOP-compatible 6D pose estimation dataset, that accurately depicts adverse challenges caused by metallic objects, i.e., reflections and specular highlights, for household items and industrial objects. Our dataset depicts 60 objects taken from Sketchfab\footnote{\url{https://sketchfab.com/}} that we subdivide into 3 categories: (a) rotational symmetric objects (\textit{Cans}) such as beer cans or energy drinks, (b) common metallic (\textit{Household}) items, such as bottles, cutlery and cups, and (c) (\textit{Industrial)} parts with both low and high level of detail. The utilized objects are illustrated in Fig. \ref{fig:objects}.

We simulate the dataset using physically-based rendering (PBR), which has shown to reduce the synthetic-to-real domain gap on several tasks \cite{51,52,53}. The images are generated using BlenderProc \cite{46} with Blender's \cite{blender} BSDF shader, simulating relevant material properties for reflective objects, i.e., \textit{metallic}, \textit{specular}, and \textit{roughness}.

Our dataset follows two strategies for object placement: \textit{Multiple Instances of the Same Object} (MiSo), where 1 to 10 instances of the same object are positioned in a scene, and \textit{Single Instance, Multiple Object} (SiMo), where we place 1 to 10 objects in a scene without duplicates of the same instance. For each object, we sample the BSDF parameters, within a reasonable range, such that they depict a wide range of visual appearance.
For each of the strategies, we render images using any combination of five lighting scenarios, which are illustrated in Fig. \ref{fig:light}, and three types of scene backgrounds: a plain black background, a plain textured with a random common floor texture taken from a subset of the CC Textures library\footnote{\url{https://cc0textures.com}}, and a real-world environment generated from a random HDR image taken from a subset of the HDR images provided by the Haven 3D Asset Library\footnote{\url{https://polyhaven.com}}.

For some of the images we introduce additional objects from the ITODD \cite{8} dataset, to have a more diverse set of objects and the dataset, and to increase the occurence frequency of occlusions. Furthermore, we introduce reflective surfaces to some to make the down-stream models more robust against environmental reflections. We illustrate these modalities in Fig. \ref{fig:scenes2}.

For each combination of object, lighting and scene background, we generate 10 \textit{MiSo} scenes, from which we render 5 camera angles each, and 120 \textit{SiMo} scenes from which we render 25 camera angles each, resulting in 45.000 MiSo and 45.000 SiMo images for our training set.
Furthermore, we render a test set containing 300 images per scene background type for MiSo and SiMo each, using background textures and HDR images, that were not used in the training set. We make sure that the test set consisting of 1.800 images contains a balanced representation of the light modalities, that were used in the training set.

\subsection{Keypoint Prediction}
Due to the usage of synthetic data generation, we have access to the ground truth geometry of the rendered objects.
We leverage this knowledge, by adding an output to the GeoHead of GDRNPP, that estimates relevant geometric features on the object surface (\textit{keypoints}).
This additional output can be controlled during the training process by adding a loss term that compares this estimate with the ground truth \textit{keypoints}.
We compare multiple strategies to incorporate these predicted \textit{keypoints} into the Patch-PnP part of GDRNPP, i.e., adding them to the other feature maps, concatenating them to the feature maps, and using a BAM  with the other feature maps.
Our results show that incorporating the prediction of relevant geometric features into the GeoHead using a BAM significantly improves the quality of the resulting 6D pose estimates.
Fig. \ref{fig:method} illustrates the overall architecture incorporating the estimated \textit{keypoints}.

\begin{figure}
    \centering
    \includegraphics[width=1\linewidth]{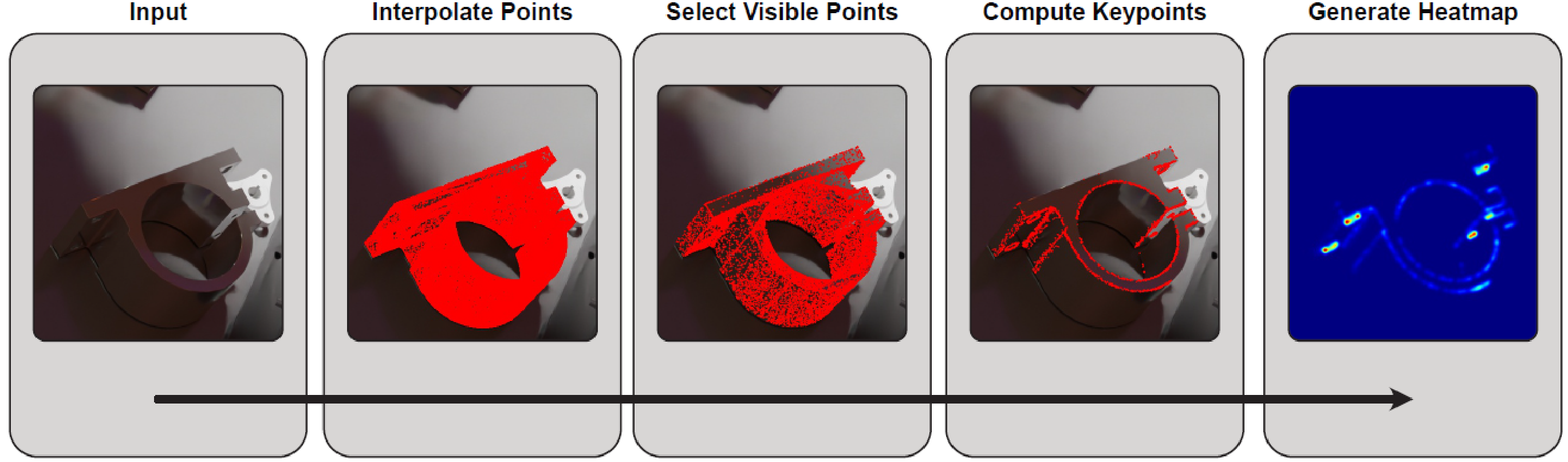}
    \caption{Given a view of the target object, we project object points into the view, remove hidden points, identify points with relevant surface information as keypoints, and finally derive a continous signal by generating a heatmap.}
    \label{fig:keypoints}
\end{figure}

The keypoint generation process is illustarted in Fig \ref{fig:keypoints}.
First, we generate a set of 3D points, that are equally distributed on the object surface, which we filter by visibility due to self-occlusion and occlusion by other objects.
We employ a saliency-based approach to identify relevant geometric features by selecting \textit{keypoints} with high surface curvature in regions with high density of potential \textit{keypoints}. We use 3D Harris features \cite{60} to get a measure for \textit{keypoint} curvature.
The algorithm calculates a covariance matrix $C_p$ for each visible point $p$ based on its nearest neighbors $\mathcal{N}(p)$:
$$
C_p = \frac{1}{|\mathcal{N}(p)|}\sum_{x_i \in \mathcal{N}(p)} (x_i - p)(x_i - p)^T .
$$
The eigenvalues $\lambda_1 \geq \lambda_2 \geq \lambda_3$ of $C_p$ represent the curvature in different directions.
Furthermore, we calculate a local density $\rho$ of potential \textit{keypoints} surrounding $p$.
Points with a saliency value $S$ given by
$$
S = \rho \frac{\lambda_1}{\lambda_1 + \lambda_2 + \lambda_3},
$$
over a certain threshold $\tau$ are considered relevant geometric feature points, i.e. \textit{keypoints}. Preliminary experiments show, that the weighting by density improves the downstream task performance in comparison to using only Harris 3D features.

Often deep learning strategies have difficulties estimating discrete feature points accurately, i.e., they usually exhibit a certain margin of error.
Furthermore, the exact position of predicted \textit{keypoints} gets lost due to the scaling difference between ground truth image and predicted output.
Therefore, we encode the \textit{keypoint} density in a weighted heatmap.
For this purpose, we project the 3D \textit{keypoints} into the image space, giving the corresponding pixel in the heatmap a value between 0 and 1, corresponding to the normalized saliency value of the 3D \textit{keypoint}.
To mitigate the effects of slight spatial errors during the estimation process, we employ a Gaussian kernel to the heatmap, distributing the weight of a \textit{keypoint} to its neighboring pixels, resulting in a more continuous representation.

\subsection{Material Properties Prediction}
\begin{figure}[t]
    \centering
    \begin{subfigure}{0.32\linewidth}
        \centering
        \includegraphics[width=\linewidth]{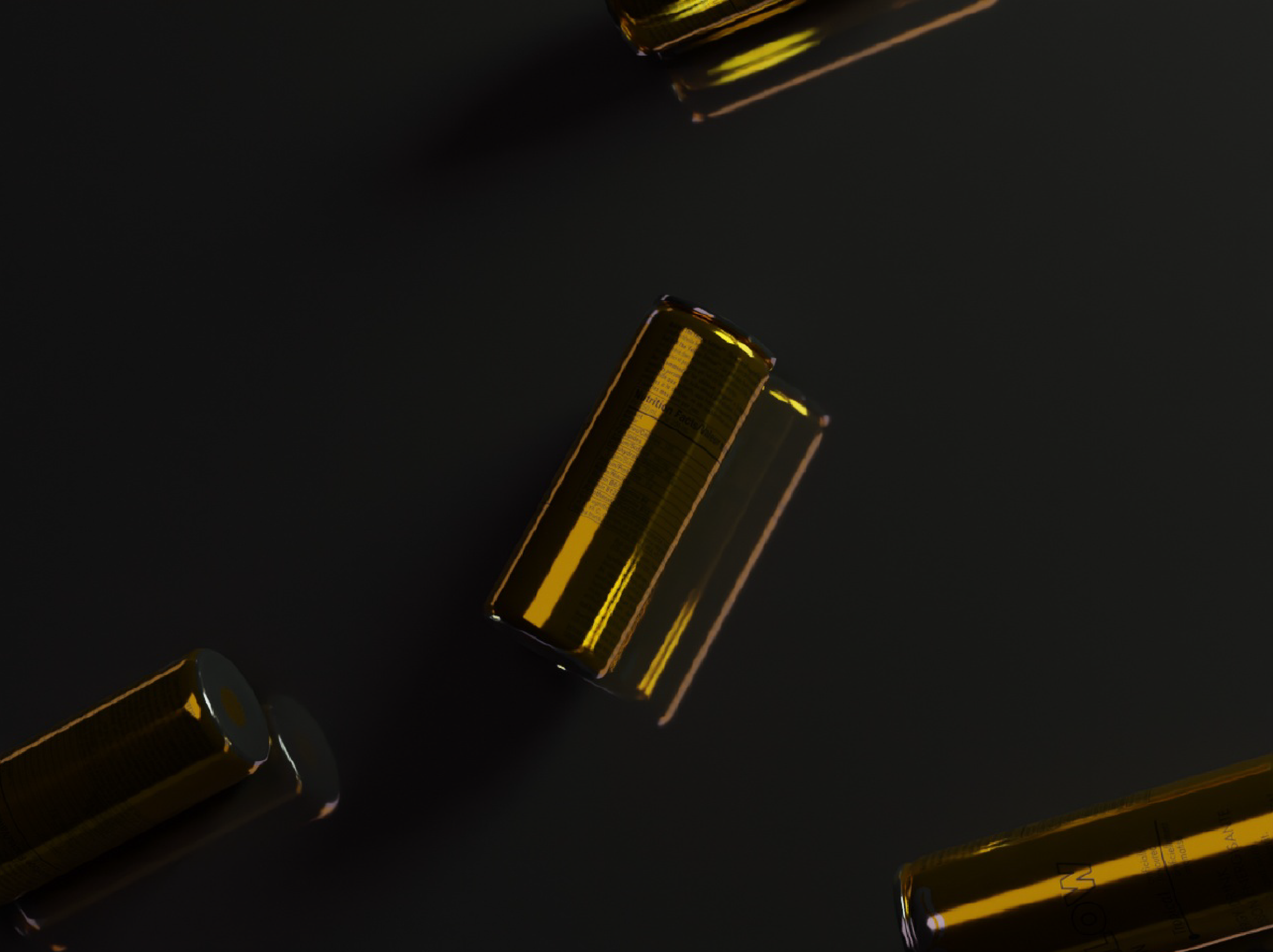}
        \caption{Original}
    \end{subfigure}
    \begin{subfigure}{0.32\linewidth}
        \centering
        \includegraphics[width=\linewidth]{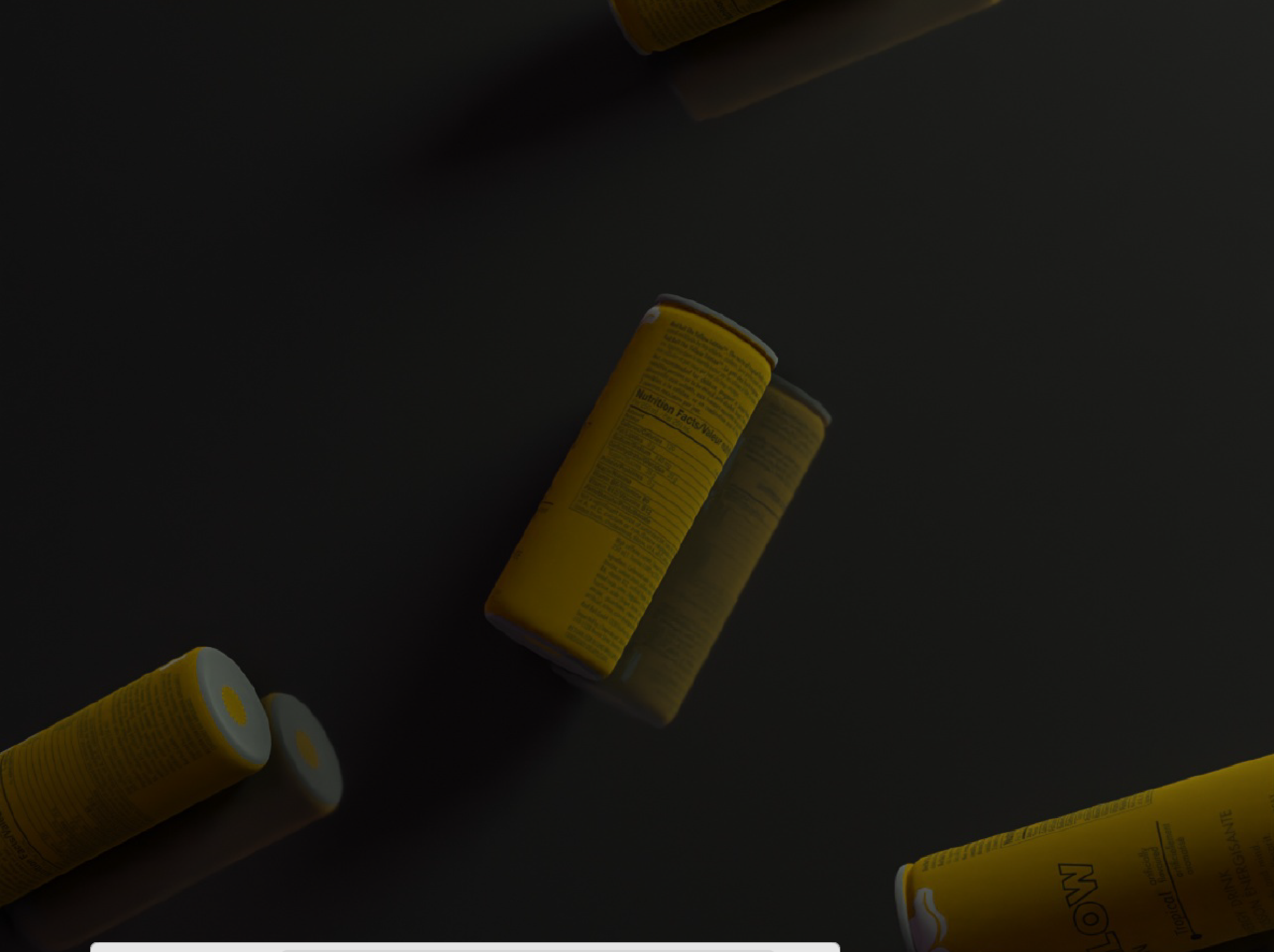}
        \caption{High Roughness}
    \end{subfigure}
    \begin{subfigure}{0.32\linewidth}
        \centering
        \includegraphics[width=\linewidth]{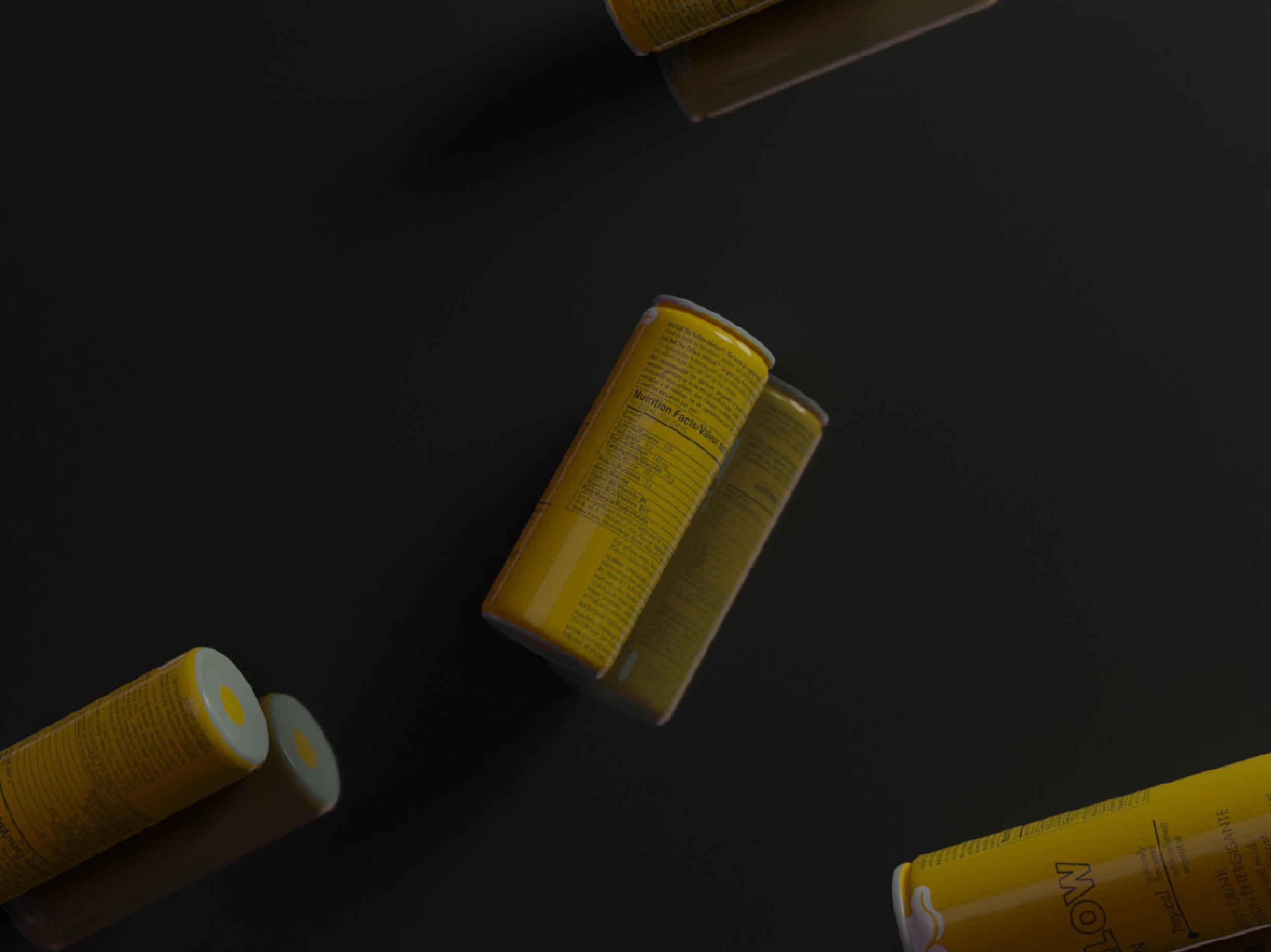}
        \caption{No Metallic}
    \end{subfigure}

    \vspace{0.1cm} 

    \begin{subfigure}{0.33\linewidth}
        \centering
        \includegraphics[width=\linewidth]{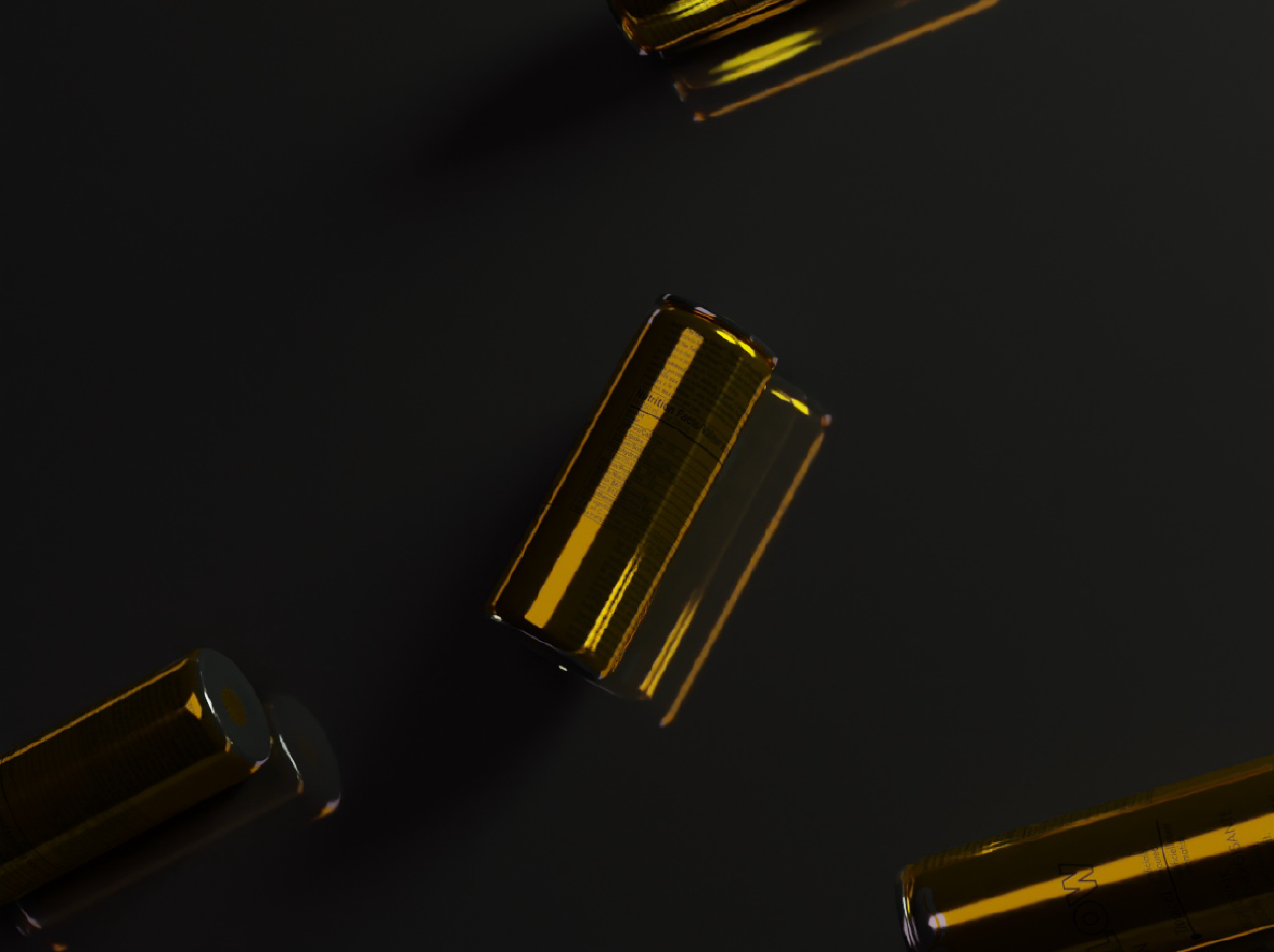}
        \caption{No Specular}
    \end{subfigure}
    \begin{subfigure}{0.33\linewidth}
        \centering
        \includegraphics[width=\linewidth]{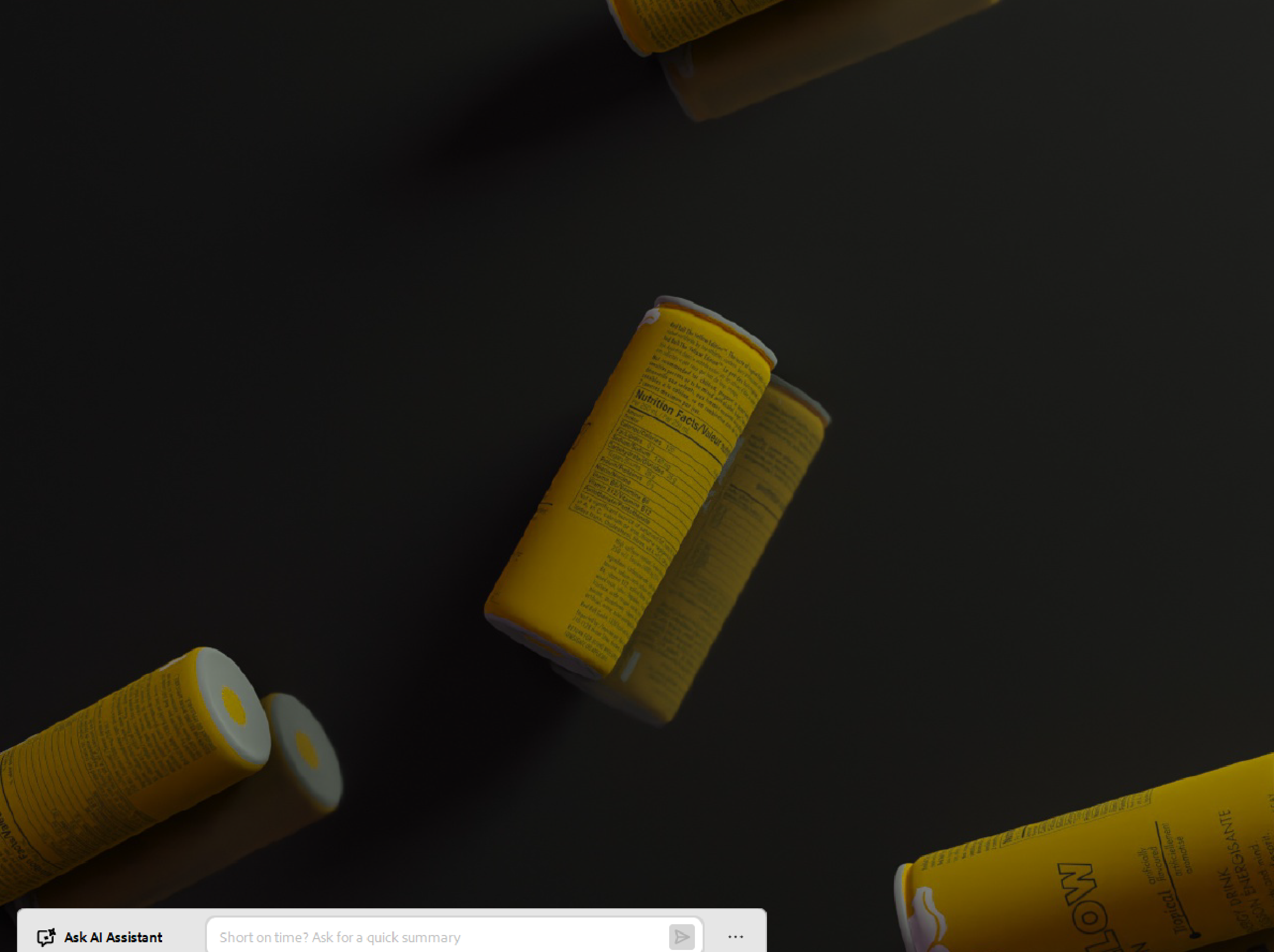}
        \caption{Combined}
    \end{subfigure}
    
    \caption{Renderings of a scene with varying material properties: (a) the original image from our dataset, (b) a version with high roughness, (c) a version without metallic properties, (d) a version without specular reflections, and (e) a combination of the properties of (b-d), being better suited for downstream tasks.}
    \label{fig:materials}
\end{figure}

6D pose estimators struggle with metallic objects due to reflections and specular highlights. However, these challenges are significantly reduced for objects with favorable material properties, such as high \textit{roughness} and low \textit{specular} and \textit{metallic} values in their BSDF representation.
Fig. \ref{fig:materials} illustrates the influence of different material properties on the rendered images.
Therefore, we propose a strategy, in which the GeoHead found in GDRNPP aims to estimate the appearance of a given object with these beneficial material properties.
Since our dataset is generated synthetically, we can easily produce images, with the same scene parameters, i.e., object placement, lighting and scene background, only differing in the BSDF parameters of the object.
We employ this second image as ground truth by adding a loss function, that enforces the additional head of the GeoHead network to predict this image with beneficial object properties.
We propose three different strategies to integrate this information into the network:
First, we directly train the GeoHead, such that the additional head outputs the image with beneficial material properties akin to an \textit{auto-encoder}. Second, we learn the \textit{difference} between the images with and without additional material properties. Finally, we use the \textit{reconstructed} image with beneficial material properties from the first strategy as input to a second forward pass of the GeoHead in GDRNPP. This essentially corresponds to a generation of the image with more suited material properties alleviating the specific issues of metallic surfaces, which leads to improved pose estimation results.


\begin{table*}[t]
    \centering
    \renewcommand{\arraystretch}{1.2}
    \begin{tabular}{ll|ccccccc}
        \toprule
        Category & Extension & AR $\uparrow$ & AD (0.1) $\uparrow$ & MSPD $\uparrow$ & MSSD $\uparrow$ & reS (10) $\downarrow$ & teS (10) $\downarrow$ & VSD $\uparrow$ \\
        \midrule
        \multirow{4}{*}{Can} 
        & GDRNPP \cite{liuShanicelGdrnpp_bop20222024}  & 30.66 & 59.72 & 19.26 & 24.27 & 3.72  & \textbf{79.76} & 48.45 \\
        & Ours (ADD)       & 27.91 & 59.20 & 19.04 & 22.40 & 4.31  & 82.46 & 42.27 \\
        & Ours (CON)       & 27.31 & 59.84 & 17.92 & 21.04 & \textbf{2.80}  & 82.40 & 42.97 \\
        & Ours (BAM)       & \textbf{33.20} & \textbf{66.58} & \textbf{22.40} & \textbf{27.52} & 5.10  & 83.53 & \textbf{49.68} \\
        \midrule
        \multirow{4}{*}{Household} 
        & GDRNPP \cite{liuShanicelGdrnpp_bop20222024}  & 17.59 & 20.67 & 19.13 & 17.91 & 1.14  & 66.47 & \textbf{15.72} \\
        & Ours (ADD)       & 18.29 & 19.68 & 24.14 & 19.27 & 2.11  & \textbf{66.00} & 11.45 \\
        & Ours (CON)       & 18.00 & 20.16 & 22.36 & 18.30 & \textbf{1.13}  & 66.14 & 13.32 \\
        & Ours (BAM)       & \textbf{23.20} & \textbf{25.64} & \textbf{29.87} & \textbf{24.46} & 3.61  & 68.98 & 15.27 \\
        \midrule
        \multirow{4}{*}{Industry} 
        & GDRNPP \cite{liuShanicelGdrnpp_bop20222024}  & 6.97  & 13.94 & 7.77  & 3.50  & \textbf{0.30}  & 71.34 & 9.63  \\
        & Ours (ADD)       & 9.32  & 11.67 & 12.73 & 5.82  & 0.81  & \textbf{70.99} & 9.40  \\
        & Ours (CON)       & 9.00  & 11.87 & 12.12 & 5.27  & 0.72  & 71.44 & 9.59  \\
        & Ours (BAM)       & \textbf{14.24} & \textbf{17.40} & \textbf{17.96} & \textbf{10.32} & 1.99  & 73.01 & \textbf{14.45} \\
        \bottomrule
    \end{tabular}
    \caption{Quantitative results of our the proposed \textit{keypoint}-based approach on the three object categories of our dataset. \textit{ADD}, \textit{CON}, and \textit{BAM} refer to the integration of the keypoint heatmap into the network, refering to addition, concatenation and integration via BAM. Best results per metric and object category are indicated in \textbf{bold}. $\uparrow$ indicates a metric, where higher values are better, $\downarrow$ indicates metrics, where lower values are better.}
    \label{tab:results_keypoints}
\end{table*}

\begin{table*}[t]
    \centering
    \renewcommand{\arraystretch}{1.2}
    \begin{tabular}{ll|ccccccc}
        \toprule
        Category & Extension & AR $\uparrow$ & AD (0.1) $\uparrow$ & MSPD $\uparrow$ & MSSD $\uparrow$ & reS (10) $\downarrow$ & teS (10) $\downarrow$ & VSD $\uparrow$ \\
        \midrule
        \multirow{4}{*}{Can} 
        & GDRNPP \cite{liuShanicelGdrnpp_bop20222024}      & 31.78 & 60.68 & 24.91 & 27.61 & 5.60  & 79.90 & 42.83 \\
        & Auto-Encoder      & 21.32 & 44.90 & 16.44 & 15.54 & \textbf{2.64}  & 73.69 & 31.99 \\
        & Difference    & 22.55 & 46.55 & 16.03 & 16.15 & 2.72  & \textbf{72.44} & 35.46 \\
        & Reconstructed & \textbf{40.81} & \textbf{63.91} & \textbf{36.56} & \textbf{36.94} & 15.08 & 80.41 & \textbf{48.92} \\
        \midrule
        \multirow{4}{*}{Household} 
        & GDRNPP \cite{liuShanicelGdrnpp_bop20222024}      & 20.56 & 23.95 & 25.85 & 21.77 & 2.13  & 66.11 & 14.05 \\
        & Ours (Auto-Encoder)      & 11.13 & 11.27 & 13.79 & 10.46 & \textbf{1.35}  & \textbf{51.41} & 8.96  \\
        & Ours (Difference)    & 15.09 & 14.70 & 20.70 & 14.16 & 1.50  & 53.96 & 10.40 \\
        & Ours (Reconstructed) & \textbf{25.31} & \textbf{27.87} & \textbf{31.22} & \textbf{26.35} & 5.73  & 66.62 & \textbf{18.35} \\
        \midrule
        \multirow{4}{*}{Industry} 
        & GDRNPP \cite{liuShanicelGdrnpp_bop20222024}      & 16.63 & 21.04 & 20.43 & 13.03 & 3.41  & 72.00 & 16.43 \\
        & Ours (Auto-Encoder)      & 6.45  & 8.34  & 9.39  & 3.01  & \textbf{0.49}  & \textbf{62.36} & 6.96  \\
        & Ours (Difference)    & 10.06 & 11.19 & 14.37 & 5.73  & 1.24  & 64.00 & 10.08 \\
        & Ours (Reconstructed) & \textbf{20.92} & \textbf{25.63} & \textbf{25.16} & \textbf{16.95} & 5.16  & 71.46 & \textbf{20.64} \\
        \bottomrule
    \end{tabular}
    \caption{Quantitative results of the proposed approach, that estimates images with beneficial material properties. Best results per metric and object category are indicated in \textbf{bold}. $\uparrow$ indicates a metric, where higher values are better, $\downarrow$ indicates metrics, where lower values are better.}
    \label{tab:results_materials}
\end{table*}

\section{RESULTS}

We evaluate the proposed methods on our dataset, training with the training split and evaluating on the test split. We follow the evaluation methodology of the BOP challenge \cite{32,hodan2024bop}, employing different metrics, that capture a broad variety of requirements of 6D pose estimation, i.e., \textit{Visible Surface Discrepancy} (VSD), \textit{Maximum Symmetry-Aware Surface Distance} (MSSD), \textit{Maximum Symmetry-Aware Projection Distance} (MSPD), \textit{Average Recall} (AR), where AR is the average of VSD, MSSD, and MSPD. In addition we report \textit{Average Distance} (AD), \textit{Rotation Symmetry Error} (reS) and \textit{Translation Symmetry Error} (teS).

\subsection{Keypoint Estimation}
The quantitative results of our \textit{keypoint}-based are illustrated in Tab. \ref{tab:results_keypoints}.
We report the results per object category comparing multiple strategies of integrating the heatmap features into the input of the Patch-PnP network with GDRNPP without extensions.
Our results show, that estimating the relevant \textit{keypoints} and incorporating them into the feature map via BAM improves the quality of 6D pose estimates over all object categories and most metrics significantly.
Our BAM-based approach effectively aligns objects with their 3D models, achieving strong performance on metrics like ADD, VSD, MSSD, and MSPD, which emphasize geometric and visual accuracy. Its slightly reduced performance in teS and reS, which demand precise absolute translation and rotation estimates, suggests a minor increase in susceptibility to depth ambiguities, object symmetries, or training biases favoring alignment over exact pose parameters. Overall, our BAM-based approach significantly outperforms its baseline, GDRNPP.

\subsection{Ideal Material Estimation}
Tab. \ref{tab:results_materials} illustrates the quantitative results of our second proposed approach, which estimates a representation of the scene with beneficial object material parameters.
These results indicate, that reconstructing the input view and requiring the network to produce a representation of more ideal material properties does not lead to an increase in performance (Auto-Encoder). Instead predicting a difference image, which depicts the pixel-wise difference between input image and a version of the input image with ideal material properties works noticeably better than the Auto-Encoder approach, but still falls short compared to the baseline. Our final approach, adding the difference image to the input image and running a second forward pass of GeoHead on this newly constructed image brings significant outperformance compared to the baseline algorithm GDRNPP. The increase in estimation quality extends across all object categories and a wide variety of metrics.

\section{CONCLUSIONS}

This work extends the GDRNPP algorithm to improve 6D pose estimation accuracy for metallic objects under challenging conditions, including reflections and specular highlights. Our new BOP-compatible dataset depicts 60 different metallic objects and introduces complex lighting scenarios, making our dataset more demanding than existing benchmarks. The proposed \textit{keypoint} learning and material parameter reconstruction both enhance performance by approximately 25\%, with the maximum gain in performance achieved on objects with complex shapes. Future work could combine these approaches to further improve accuracy. Furthermore, the introduction of additional iterative refinements could improve the trailing results of our BAM-based approach in comparison to GDRNPP regarding the rotation and translation symmetry error.

Our BOP compatible dataset is available at \href{https://huggingface.co/datasets/tpoellabauer/IGD/}{HuggingFace}.











\printbibliography

\addtolength{\textheight}{-12cm}   


\end{document}